\documentclass[conference]{IEEEtran}
\IEEEoverridecommandlockouts
\usepackage{cite}
\usepackage{amsmath,amssymb,amsfonts}
\usepackage{algorithmic}
\usepackage{graphicx}
\usepackage{textcomp}
\usepackage{xcolor}
\usepackage{url}
\def\BibTeX{{\rm B\kern-.05em{\sc i\kern-.025em b}\kern-.08em
    T\kern-.1667em\lower.7ex\hbox{E}\kern-.125emX}}
\begin{document}

\title{A Contextual Analysis of Driver-Facing and Dual-View Video Inputs for Distraction Detection in Naturalistic Driving Environments}

\author{\IEEEauthorblockN{Anthony Dontoh}
\IEEEauthorblockA{\textit{Dept. of Civil, Construction, and} \\
\textit{Environmental Engineering}\\
The University of Memphis\\
Memphis, TN, USA\\
Email: adontoh@memphis.edu}
\and
\IEEEauthorblockN{Stephanie Ivey, Ph.D.}
\IEEEauthorblockA{\textit{Dept. of Civil, Construction, and} \\
\textit{Environmental Engineering}\\
The University of Memphis\\
Memphis, TN, USA\\
Email: ssalyers@memphis.edu}
\and
\IEEEauthorblockN{Armstrong Aboah, Ph.D.}
\IEEEauthorblockA{\textit{Dept. of Civil, Construction, and} \\
\textit{Environmental Engineering}\\
North Dakota State University\\
Fargo, ND, USA\\
Email: armstrong.aboah@ndsu.edu}
}

\maketitle

\begin{abstract}
Despite increasing interest in computer vision-based distracted driving detection, most existing models rely exclusively on driver-facing views and overlook crucial environmental context that influences driving behavior. This study investigates whether incorporating road-facing views alongside driver-facing footage improves distraction detection accuracy in naturalistic driving conditions. Using synchronized dual-camera recordings from real-world driving, we benchmark three leading spatiotemporal action recognition architectures: SlowFast-R50, X3D-M, and SlowOnly-R50. Each model is evaluated under two input configurations: driver-only and stacked dual-view. Results show that while contextual inputs can improve detection in certain models, performance gains depend strongly on the underlying architecture. The single-pathway SlowOnly model achieved a 9.8 percent improvement with dual-view inputs, while the dual-pathway SlowFast model experienced a 7.2 percent drop in accuracy due to representational conflicts. These findings suggest that simply adding visual context is not sufficient and may lead to interference unless the architecture is specifically designed to support multi-view integration. This study presents one of the first systematic comparisons of single- and dual-view distraction detection models using naturalistic driving data and underscores the importance of fusion-aware design for future multimodal driver monitoring systems.
\end{abstract}


\section{Introduction}
Distracted driving remains a major threat to transportation safety, contributing to over 3,000 fatalities and 289,000 injuries annually in the U.S., accounting for about 8\% of all traffic deaths~\cite{cdc2023,nhtsa2023}. International statistics show similarly troubling patterns, with distraction implicated in 16\% of road deaths in Australia and over 12\% in Norway~\cite{who2022}. The economic burden in the U.S. exceeds \$98 billion annually~\cite{nhtsa2023economic}, fueled by increasing mobile device usage and in-vehicle technologies. These alarming trends have prompted researchers to develop intelligent monitoring systems ~\cite{kyem2025task} to detect distractions in real time.

While earlier research relied on crash reports and retrospective investigations, recent work increasingly uses naturalistic driving data~\cite{shrp2015dataset} that include synchronized video and sensor recordings from real-world driving. These datasets enable deeper behavioral insights by capturing both internal and external vehicle conditions~\cite{yan2022multi,koay2022detecting}. 

Several recent studies have leveraged naturalistic driving data to build detection algorithms. Yadawadkar et al.~\cite{yadawadkar2018identifying} used head pose and steering features to identify distracted behavior with high F1 scores. Liu et al.~\cite{liu2021identification} analyzed gaze and hand cues, while Martin et al.~\cite{martin2019driveact} used multimodal data to model upper-body and facial features. Peruski et al.~\cite{peruski2024exploring} applied deep learning to multi-camera footage but still focused only on driver-facing attributes like eye closure and hand movement.

Despite these advancements, most studies are limited by their narrow reliance on in-cabin features. For example, when a driver turns their head or glances away from the road, current systems may misclassify this as distraction, even if it reflects appropriate situational awareness. This includes actions such as checking mirrors, observing pedestrians, or responding to hazards. Without road-facing context, models cannot distinguish purposeful behavior from actual distractions like texting or personal grooming.

This limitation presents a major challenge: high false positive rates that degrade system trust and reliability. Dontoh et al.~\cite{dontoh2025visual} showed that while driver-only models perform well on curated datasets, they struggle in naturalistic settings precisely because they lack environmental awareness. Although multi-camera data is increasingly available, few systems incorporate road-facing inputs, which means they miss crucial context. Leveraging such information could help disambiguate attention-related behaviors and improve classification accuracy in real-world driving.

These findings motivate a key research question: can combining driver-facing and road-facing video inputs significantly improve distraction detection in naturalistic environments? This study addresses that gap by evaluating whether dual-view video leads to measurable performance gains over traditional driver-only models using deep learning on real-world driving data.

The primary contributions of this study are:
\begin{itemize}
  \item \textit{Systematic Evaluation of Multi-View Integration}: We provide one of the first controlled comparisons between single-view and dual-view distraction detection models using naturalistic data.
  \item \textit{Real-World Performance Assessment}: We evaluate three state-of-the-art action recognition architectures on real-world video to assess generalizability and robustness.
  \item \textit{Design Insights for Context-Aware Systems}: We show that naive input stacking offers limited gains and highlight the need for fusion-aware architectures that properly leverage contextual cues.
\end{itemize}

The remainder of the paper is organized as follows: Section II reviews related literature. Section III outlines the dataset, models, and experimental setup. Section IV presents results and analysis. Section V concludes with key findings and future directions.

\section{Related Work}
Research on distracted driving detection using naturalistic data has evolved significantly with the availability of large-scale datasets and advances in computer vision techniques. This section reviews key developments in two main areas: the naturalistic driving datasets that have enabled this research, and the predominant single-view approaches that have been developed using these datasets.

\subsection{Naturalistic Driving Datasets}

The availability of large-scale naturalistic driving datasets has transformed distraction detection research by providing real-world, multi-perspective video and sensor data. The SHRP2 study~\cite{shrp2015dataset} remains foundational, offering synchronized in-vehicle footage and kinematic data across diverse U.S. drivers. Subsequent datasets such as DMD~\cite{ortega2020dmd}, Drive\&Act~\cite{martin2019driveact}
, and 3MDAD~\cite{yan2022multi} have expanded the field with multimodal annotations, multi-angle views, and fine-grained behavioral labels. Benchmarks like SFDDD~\cite{sfddd2016} have supported early classification tasks, while black box recordings and commercial dashboard cameras now offer naturalistic video under variable lighting, weather, and traffic conditions. Urban datasets like Cityscapes~\cite{cordts2016cityscapes} and KITTI~\cite{geiger2013kitti}, though not distraction-focused, provide contextual environmental data for road scene understanding.

Despite these advances, most research still emphasizes driver-facing views, underutilizing road-facing context that could enhance detection of situational awareness versus actual distraction.

\subsection{Single-View Detection Approaches}

Most distracted driving models rely on driver-facing views to analyze head pose, gaze, and upper-body cues. Early studies used hand-crafted features with traditional classifiers, while recent works have adopted deep learning. CNN-based methods and hybrid CNN-RNN architectures have shown promise, with some models integrating attention mechanisms~\cite{danyo2025improved,he2025rmtse}
to improve focus.

However, single-view systems often struggle in real-world scenarios. Ezzouhri et al.~\cite{ezzouhri2021robust} addressed robustness issues, but contextual limitations remain. Liu et al.~\cite{liu2021identification} and Koay et al.~\cite{koay2022detecting} found that without road context, such models misclassify behaviors like head turns meant for situational awareness. Benchmarks like SFDDD~\cite{sfddd2016} further limit generalizability due to their controlled nature.

To address this, we compare dual-view and single-view inputs using naturalistic driving data to assess whether road context improves distraction detection accuracy.

\section{Methodology}
\begin{figure}[htbp]
\centerline{\includegraphics[width=0.95\linewidth]{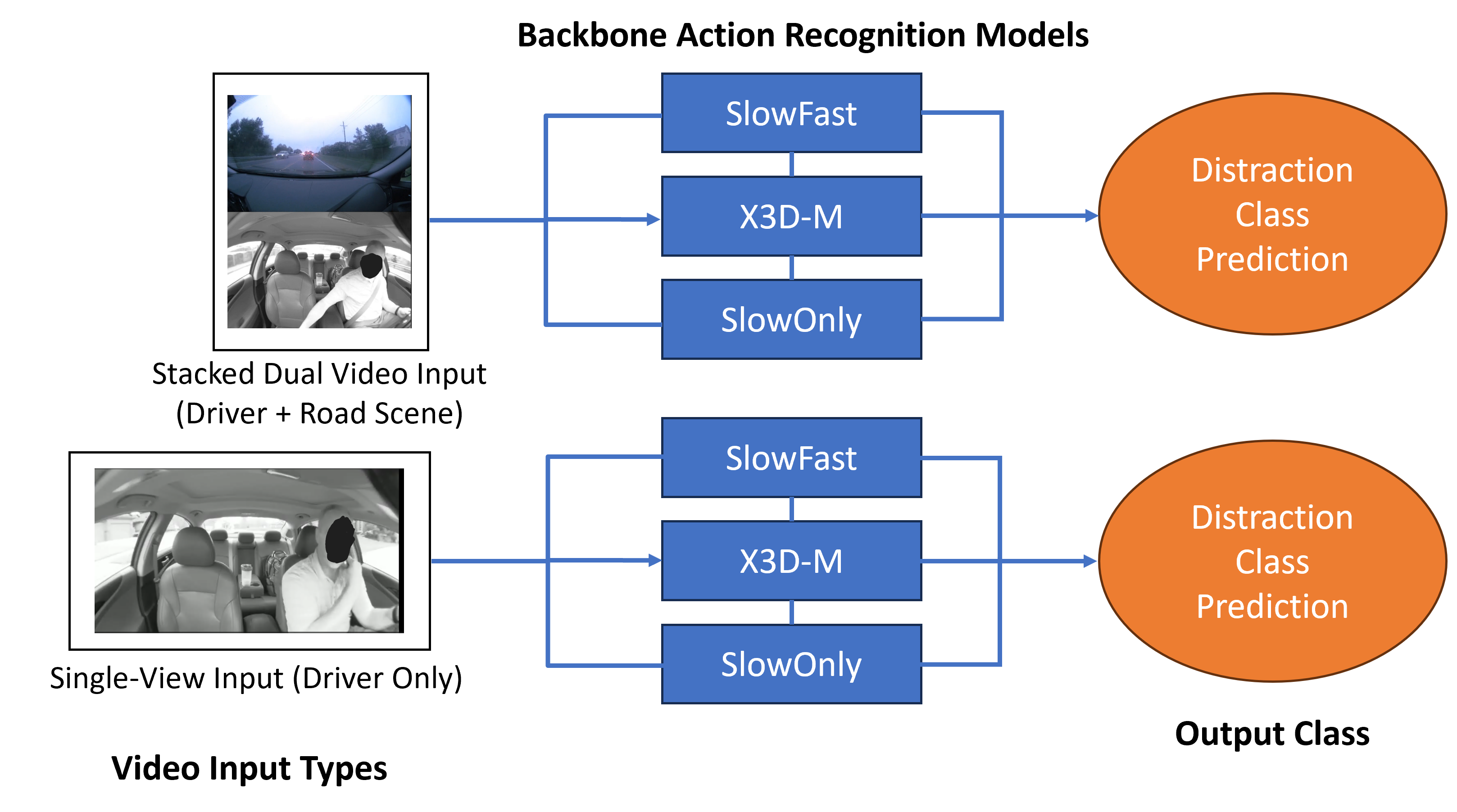}}
\caption{Architectural block diagram of the experimental setup. Both dual-view (driver + road) and single-view (driver only) inputs are evaluated using three backbone action recognition models (SlowFast, X3D-M, and SlowOnly), with outputs classified into distraction categories.}
\label{fig:method_architecture}
\end{figure}

\noindent
The experimental methodology employed in this study follows a systematic comparative framework designed to isolate the effects of environmental context on distracted driving detection performance. Figure~\ref{fig:method_architecture} presents the \textit{architectural block diagram} of the experimental setup, illustrating the two input configurations evaluated under controlled conditions: a multimodal dual-view configuration that integrates synchronized driver-facing and road-facing video streams to capture both behavioral and environmental cues, and a traditional single-view configuration utilizing solely driver-facing footage for baseline comparison. Each input modality is independently processed through three state-of-the-art spatiotemporal architectures: SlowFast-R50, X3D-M, and SlowOnly-R50, which have been adapted from their original action recognition frameworks to perform six-class distraction classification. This methodological approach enables rigorous performance assessment across input modalities while maintaining architectural consistency and standardized training protocols, ensuring that observed performance differences can be attributed specifically to the inclusion or exclusion of environmental context rather than variations in model implementation or experimental conditions.

\subsection{Dataset Description and Preparation}
This study utilizes a dataset derived from naturalistic driving recordings collected using blackbox sensors developed by Digital Artefacts LLC, as described by Drincic et al.~\cite{drincic2020blackbox}. The blackbox sensors were installed in individual personal vehicles and continuously recorded activities inside and outside the vehicle, including synchronized driver-facing and road-facing camera views. To prepare the data for model training, the continuous recordings were segmented into fixed-length clips of 5 seconds each, sampled at 25 frames per second (FPS), yielding 125 frames per clip using OpenCV and MoviePy libraries. Two dataset variants were prepared for model training to enable systematic comparison between input modalities. The first variant, termed stacked dual-view videos, retains both synchronized views as a single composite input, with the road-facing view positioned on top and the driver-facing view on the bottom. The second variant, driver-only videos, was created by cropping the lower half of each composite video to isolate only the driver-facing view, thereby simulating traditional single-camera driver monitoring systems.

Each clip was manually reviewed and labeled into one of six categories through simultaneous inspection of both driver-facing and road-facing views. This dual-view labeling process enabled annotators to assess driver behavior contextually, considering road conditions and traffic situations to distinguish between legitimate situational awareness behaviors and actual distractions. The classification framework encompasses six distinct behavioral categories: hands off the wheel, characterized by visible disengagement from the steering wheel for prolonged periods; head turned, indicating drivers turning their head away from the road and suggesting loss of situational awareness; listening to radio or music, capturing interactions with audio systems accompanied by observable physical and attentional shifts; normal driving, representing baseline behavior where drivers maintain full attention on the road with proper hand positioning; personal grooming, including behaviors such as hair fixing, makeup application, or mirror adjustment while the vehicle is in motion; and stopped, indicating stationary vehicle conditions at lights, in traffic, or while parked, allowing the model to distinguish distraction during motion versus inactivity.

\begin{figure}[htbp]
\centerline{\includegraphics[width=0.85\linewidth]{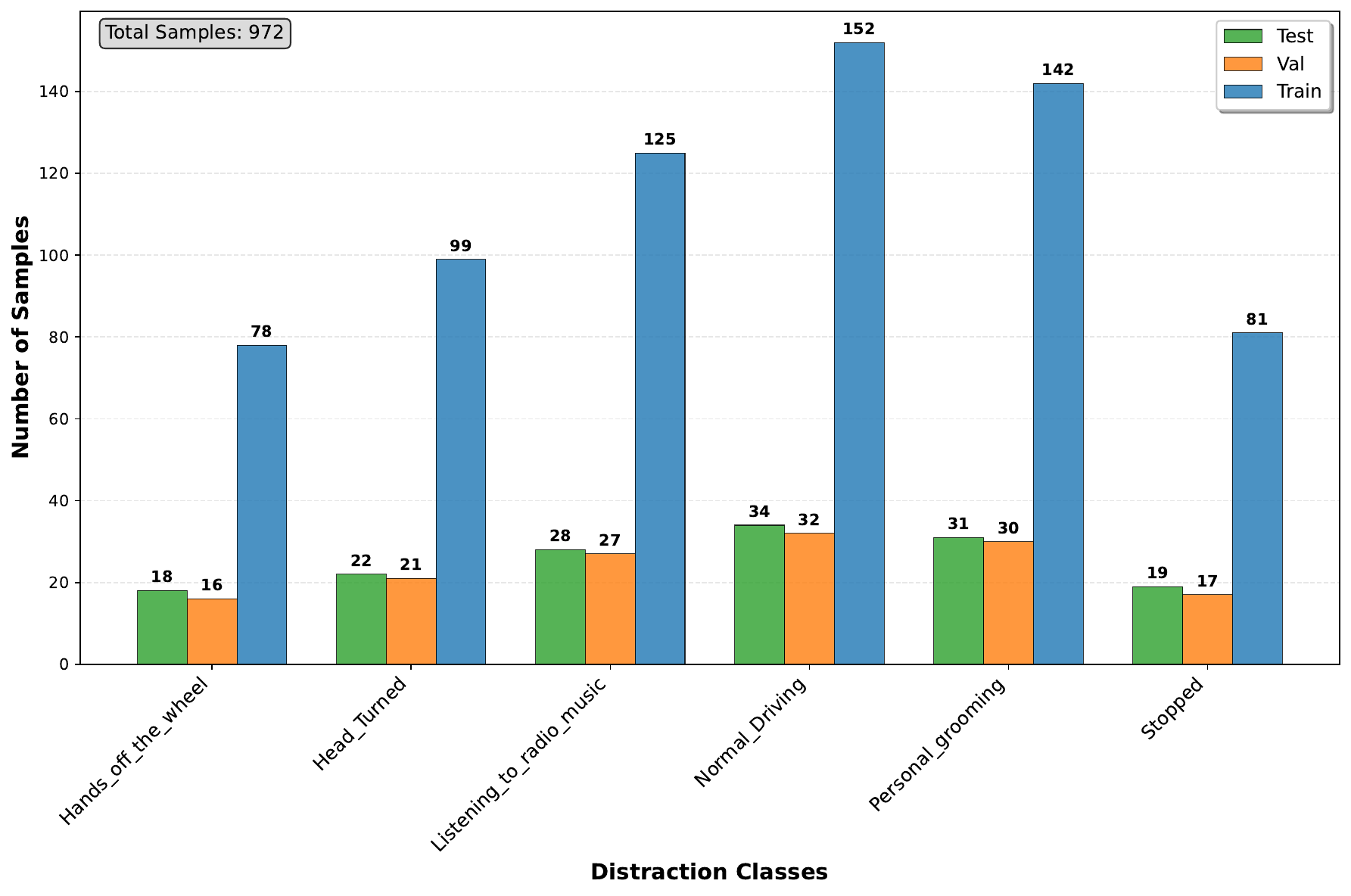}}
\caption{Distribution of distraction classes across training, validation, and test sets.}
\label{fig:class_dist}
\end{figure}

All video clips were resized to a fixed side length of 256 pixels, followed by center cropping to 224×224 pixels to match model input dimensions. The dataset was partitioned into training (70\%), validation (15\%), and test (15\%) subsets, maintaining balanced class distributions to ensure robust model evaluation and prevent overfitting. Figure~\ref{fig:class_dist} shows the distribution of samples per class across the training, validation, and test subsets.

\subsection{Model Architectures}
This study benchmarks three state-of-the-art video recognition architectures: SlowFast-R50~\cite{feichtenhofer2019slowfast}, X3D-M~\cite{feichtenhofer2020x3d}, and SlowOnly-R50~\cite{feichtenhofer2019slowfast}. All three models originate from the PyTorchVideo repository and are recognized for their strong performance on the Kinetics dataset, a benchmark for human action classification. These architectures were selected because distraction detection is fundamentally a spatiotemporal recognition problem, requiring models that can capture both visual semantics and motion patterns over time.

\subsubsection{SlowFast Networks}
SlowFast~\cite{feichtenhofer2019slowfast} is a dual-pathway architecture inspired by the primate visual system, with a slow pathway capturing spatial semantics at low frame rates and a fast pathway emphasizing motion dynamics at higher temporal resolution with reduced channels. In this study, we employ the SlowFast-8×8-R50 configuration, where the slow branch samples 8 frames at stride 8 using a ResNet-50 backbone and the fast branch processes 32 frames with 1/8 channel capacity. The pathways are fused through lateral connections at multiple stages, enabling efficient integration of spatial and temporal features while keeping the fast branch at only 20\% of the total computation.

\subsubsection{X3D Networks}

X3D~\cite{feichtenhofer2020x3d} is an efficient video recognition architecture that expands a lightweight 2D model along six axes—temporal duration ($\gamma_t$), frame rate ($\gamma_\tau$), spatial resolution ($\gamma_s$), network width ($\gamma_w$), bottleneck width ($\gamma_b$), and depth ($\gamma_d$). The X3D-M variant used in this study applies a coordinate descent strategy starting from a MobileNet-like baseline. It processes 16 frames at 224×224 resolution with expansion factors $\gamma_\tau = 5$, $\gamma_t = 16$, $\gamma_s = 2$, $\gamma_w = 1$, $\gamma_b = 2.25$, and $\gamma_d = 2.2$, resulting in 4.73 GFLOPs and 3.76M parameters. X3D-M includes Squeeze-and-Excitation blocks and Swish activations, maintaining full temporal resolution until global pooling. This allows efficient spatiotemporal modeling with minimal computational cost.

\subsubsection{SlowOnly Networks}

SlowOnly~\cite{feichtenhofer2019slowfast} represents a simplified single-pathway architecture derived from the SlowFast framework, retaining only the slow pathway while maintaining identical training procedures and hyperparameters to enable controlled comparison of multi-pathway benefits. The architecture employs a ResNet-50 backbone that processes 4 frames with temporal stride 16 at 224×224 spatial resolution, using a strategic temporal convolution placement where early layers (conv1 to res3) utilize essentially 2D convolution kernels and later layers (res4 and res5) implement non-degenerate temporal convolutions to capture motion patterns. This design choice is motivated by observations that temporal convolutions in earlier layers can degrade accuracy when objects move quickly and temporal stride is large, while sufficient spatial receptive field size in deeper layers enables effective temporal correlation modeling. Despite its architectural simplicity compared to multi-pathway approaches, SlowOnly achieves competitive performance on video recognition benchmarks, serving as both a strong baseline and a means to understand when sophisticated temporal modeling approaches provide additional benefits.

\subsection{Training Strategy and Hyperparameters}

All models were trained using the Adam optimizer with a fixed learning rate of 1e-4 and a batch size of 8. Adam was chosen for its stable convergence on small-to-moderate datasets and consistent results in preliminary trials. Training was conducted on a SLURM-managed high-performance computing cluster equipped with NVIDIA Ada 6000 and H100 GPUs. Cross-entropy loss was used as the objective function, given the categorical nature of the six distraction classes. Each model was trained for up to 50 epochs, with early stopping based on validation accuracy to mitigate overfitting. The checkpoint yielding the highest validation performance was retained for final evaluation.

To tailor each model to the six-class distraction detection task, we replaced the final fully connected classification layer with a 6-way output. All models were initialized with Kinetics-400 pre-trained weights to leverage robust spatiotemporal feature priors. Training was performed under two conditions: (1) using only the driver-facing video stream, and (2) using stacked dual-view inputs combining synchronized driver-facing and forward-facing clips. This allowed for comparative assessment of the added value of contextual scene information. Video clips were resized and normalized to meet the expected input dimensions and statistical properties of each architecture, with aspect ratios preserved to retain spatial fidelity. The selected models; SlowFast, X3D-M, and SlowOnly; represent distinct design philosophies. SlowFast incorporates a dual-pathway structure that separates slow and fast temporal dynamics. X3D-M is a depth-expandable and computationally efficient architecture. SlowOnly serves as a straightforward temporal baseline with uniform frame sampling. Together, these architectures enable a comprehensive evaluation of how different design strategies affect contextual distraction detection performance.

\subsection{Evaluation Metrics}

Model performance was evaluated using overall classification accuracy on the test set, defined as the percentage of correctly predicted samples across all distraction categories. This metric provides a general indication of how well the model distinguishes between different types of driver behavior.

To provide a more detailed assessment, confusion matrices were generated for each model and input configuration. These matrices helped identify which distraction categories were most commonly misclassified, revealing patterns of confusion between visually similar behaviors such as grooming and radio interaction.

In addition to accuracy, macro and weighted F1 scores were computed to account for class imbalance and assess per-class performance. The macro F1 score treats all classes equally, while the weighted F1 score adjusts for the frequency of each class in the dataset. Model training and validation performance were tracked over time using TensorBoard logs, allowing for visual inspection of convergence trends and the early detection of overfitting during training.

\section{Results and Discussion}
This section presents the results with discussions of the comparative analysis run in this study.

\subsection{Model Performance Overview}

\begin{figure}[htbp]
\centerline{\includegraphics[width=0.95\linewidth]{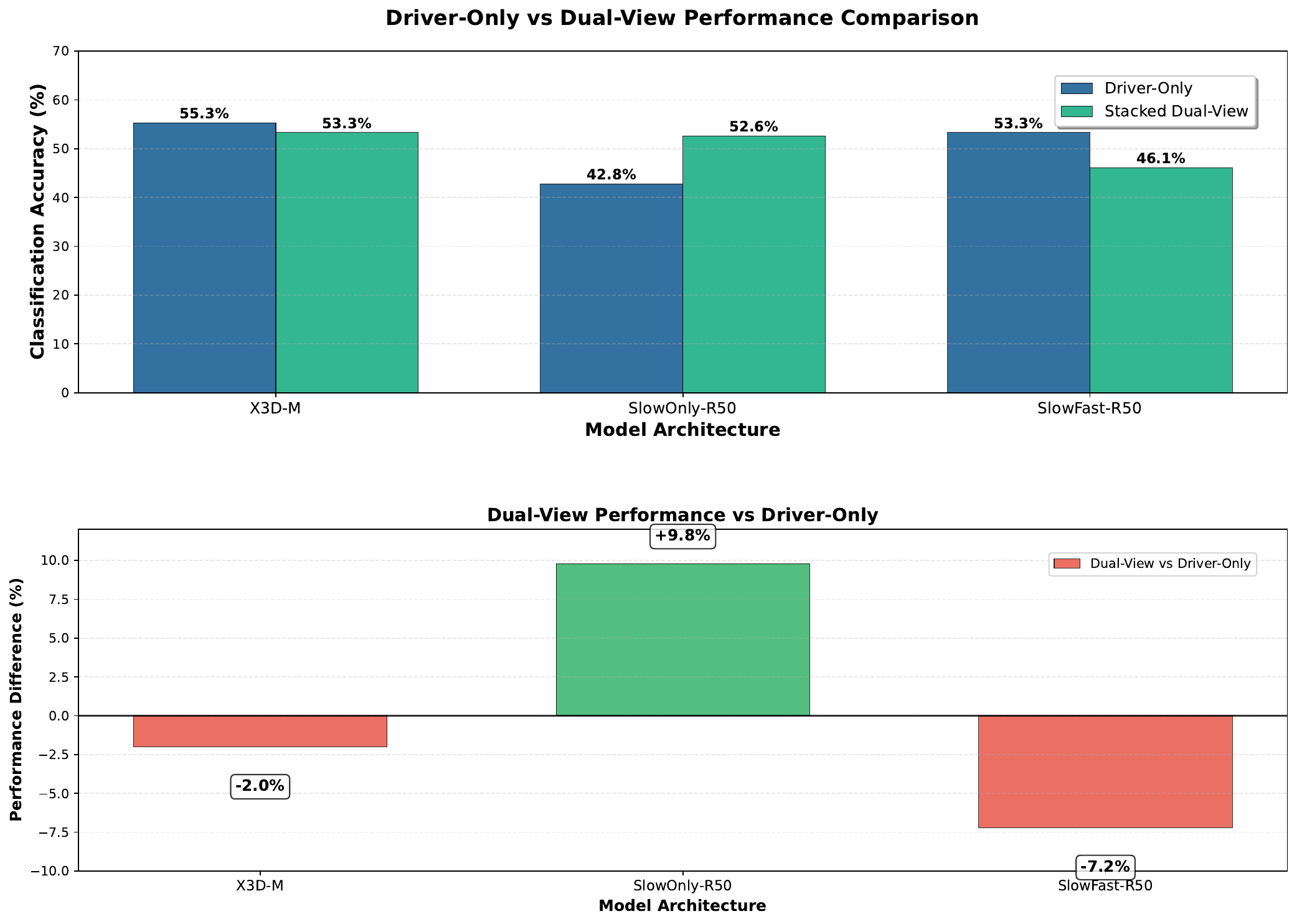}}
\caption{Comparison of classification accuracy across model architectures using driver-only versus stacked dual-view inputs (top), and corresponding accuracy differences (bottom).}
\label{fig:model_perf_comparison}
\end{figure}

To evaluate the effectiveness of incorporating road-facing context via stacked dual-view inputs, each of the three baseline architectures was trained and tested under two configurations: using only the driver-facing video stream and using synchronized driver- and road-facing streams as stacked inputs. This comparison was designed to isolate the impact of external scene context on distraction classification performance. The dual-view setup aimed to enhance contextual understanding, particularly for distraction classes that are visually ambiguous from the cabin view alone, such as head turns, reaching movements, or glances away from the road. The results of this evaluation are summarized in Figure~\ref{fig:model_perf_comparison}.

The best overall performance was achieved by the driver-only X3D-M model, with a test accuracy of 55.3\%, demonstrating the effectiveness of its progressive expansion methodology for capturing driver behavior patterns. The results reveal striking architectural differences in multi-view information processing. SlowOnly-R50 showed the most substantial improvement when provided with stacked inputs, gaining 9.8\% over its driver-only counterpart, suggesting that its single-pathway design can effectively leverage additional spatial context without architectural conflicts. This improvement indicates that the simplified temporal modeling in SlowOnly creates capacity for processing complementary visual information from the road-facing camera.

Conversely, SlowFast-R50 suffered a significant drop in accuracy (7.2\%) when road-facing input was included, revealing that its dual-pathway architecture may experience interference between the fast pathway's motion detection and external scene dynamics. The fast pathway, designed to capture fine-grained temporal changes in driver behavior, appears to be disrupted by the additional motion patterns from road scenes, potentially creating conflicting signals that degrade classification performance. X3D-M showed minimal sensitivity to input configuration (2.0\% decrease), reflecting its architectural robustness through progressive optimization across multiple network axes. These findings suggest that while stacked dual-view inputs can enhance performance in certain architectures, the benefits are highly dependent on the underlying temporal modeling strategy, with simpler single-pathway designs showing greater adaptability to multi-view fusion than complex dual-pathway systems optimized for specific motion patterns.

\subsection{Per-Class Analysis}

To further understand the behavioral distinctions learned by each model, the study analyzed per-class performance across all six model configurations. Table~\ref{tab:perclass_summary} summarizes accuracy and macro F1 score for each model-view pairing. This analysis complements the model-level overview in the previous section by illustrating how class-specific recognition varies with and without stacked video input, revealing important insights about architectural sensitivity to multi-view information fusion. The macro F1 scores provide a balanced assessment of per-class performance that accounts for class imbalances in the dataset, while weighted F1 scores reflect the natural distribution of distraction behaviors. These metrics collectively demonstrate that model performance extends beyond overall accuracy, with significant variations in how different architectures handle class-specific patterns and multi-view integration, particularly evident in the substantial differences between driver-only and dual-view configurations across the three temporal modeling approaches.

\begin{table}[!ht]
    \caption{Model Performance Comparison}\label{tab:perclass_summary}
    \begin{center}
        \begin{tabular}{l l l l l}
            Model & View & Accuracy & Macro F1 & Weighted F1 \\\hline
            X3D-M & Dual-View & 0.53 & 0.45 & 0.44 \\
            X3D-M & Driver Only & 0.55 & 0.45 & 0.45 \\
            SlowFast & Dual-View & 0.46 & 0.42 & 0.42 \\
            SlowFast & Driver Only & 0.53 & 0.51 & 0.50 \\
            SlowOnly & Dual-View & 0.53 & 0.44 & 0.44 \\
            SlowOnly & Driver Only & 0.43 & 0.41 & 0.40 \\\hline
        \end{tabular}
    \end{center}
\end{table}

\begin{figure}[htbp]
\centerline{\includegraphics[width=0.95\linewidth]{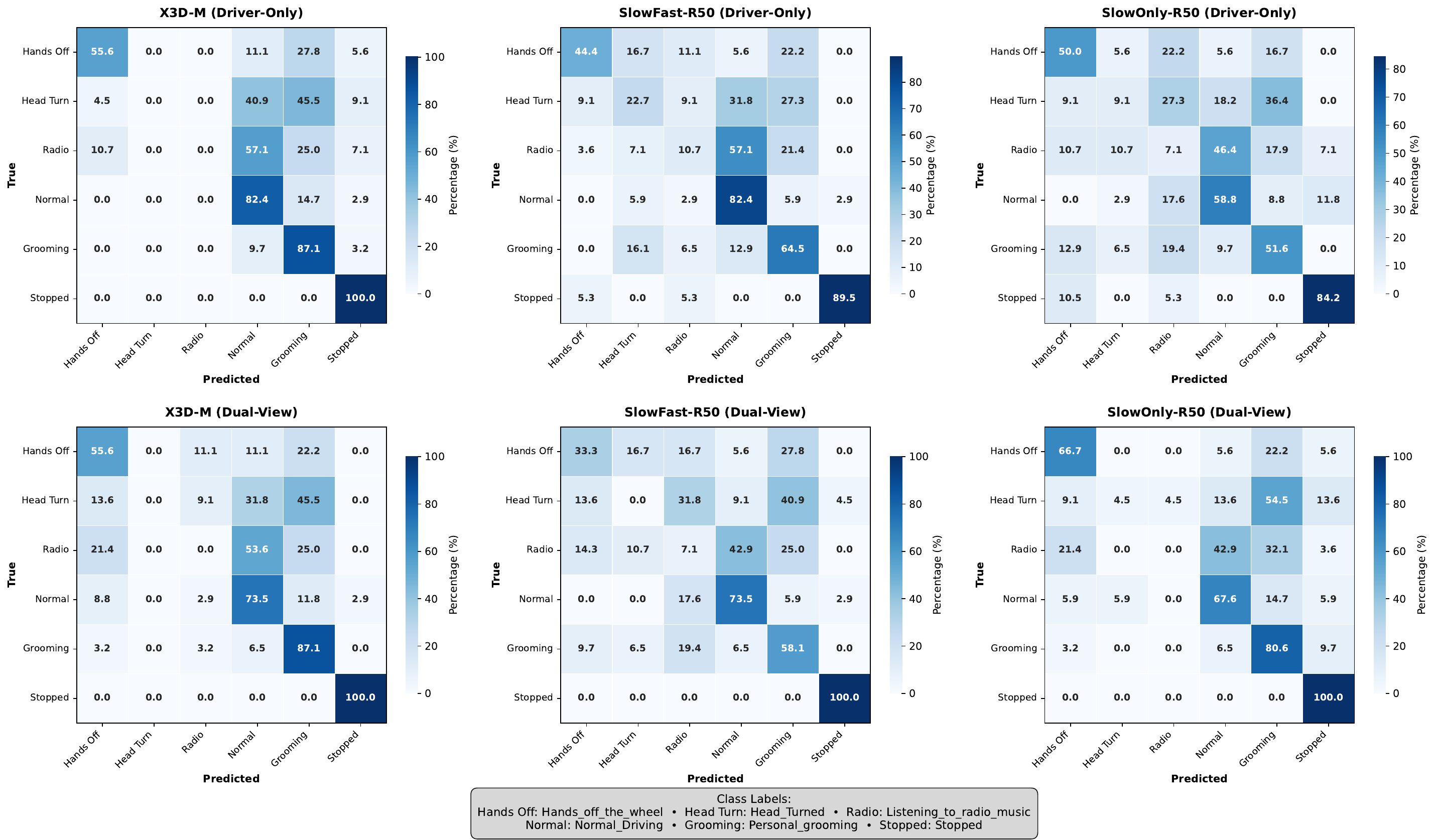}}
\caption{Confusion matrices comparing classification results across model architectures using driver-only (left) and stacked dual-view inputs (right).}
\label{fig:confusion_grids}
\end{figure}

Figure~\ref{fig:confusion_grids} presents confusion matrices for all six baseline models, grouped by architecture and input modality. These matrices reveal nuanced patterns in classification performance across distraction categories and provide insight into the relative strengths and weaknesses of single- and dual-view input configurations.

Across all models and modalities, the \textit{Stopped} class consistently achieves near-perfect accuracy, with predictions exceeding 95\% in most cases. This class is characterized by minimal driver movement and a stable visual appearance, making it trivially separable from dynamic distraction behaviors. Similarly, \textit{Normal driving} and \textit{Personal grooming} show relatively strong recognition rates, particularly for the X3D-M architecture, which appears well-suited to detecting high-frequency visual patterns associated with grooming gestures. In contrast, categories such as \textit{Head turned} and \textit{Radio interaction} exhibit high levels of confusion regardless of input modality. These behaviors are often subtle and temporally brief, involving quick glances or gaze shifts that current models cannot reliably capture. For instance, \textit{Head turned} is frequently mistaken for grooming or interacting with the radio, highlighting the need for incorporating explicit gaze modeling in future systems.

It is also observed from the results that the addition of road-facing context via dual-view inputs does not universally improve model performance. In some cases, such as SlowFast-R50, dual-view input even introduces instability, reducing classification accuracy for \textit{Grooming} and \textit{Hands off the wheel}. This performance degradation suggests that simply stacking inputs from two cameras may lead to representational noise or confusion, especially if the model lacks a dedicated fusion mechanism to differentiate salient features from each view. Rather than providing complementary context, the added visual stream can act as a distractor if not meaningfully integrated. The SlowOnly architecture, on the other hand, demonstrates modest improvements in classifying \textit{Head turned} and \textit{Radio} when using dual-view inputs. This could be attributed to its more consistent frame sampling strategy, which may benefit from the added spatial information without becoming overwhelmed by motion discrepancies between the two views. Nonetheless, even in this case, the gains are marginal, highlighting a key limitation of early fusion strategies where inputs are simply concatenated or stacked without architectural adaptation.

These findings underscore that simply stacking driver- and road-facing inputs does not reliably improve distraction detection performance. The drop observed in SlowFast-R50 suggests that dual-pathway architectures optimized for motion dynamics can be disrupted when scene motion is introduced without explicit fusion. Stacked inputs may create conflicts where features from one view overshadow the other, while the modest gains in SlowOnly indicate simpler models may adapt better, though improvements remain limited. 

Unlocking the potential of dual- or multi-view systems will require more than naive concatenation. Promising strategies include processing each view separately with mid- or late-level fusion, applying attention to prioritize salient cues, and aligning temporal signals across views. With these design considerations, dual-view architectures could reduce false positives and achieve the contextual awareness needed for reliable real-world distraction detection.

\section{Conclusion}

This study examined whether adding road-facing context through dual-view video inputs improves distraction detection compared to using driver-facing views alone. Using naturalistic driving data and three state-of-the-art video classification models, we conducted a systematic evaluation across multiple architectures and input setups.

Findings revealed that dual-view inputs do not guarantee improved performance. While the SlowOnly-R50 model showed a 9.8\% gain with dual-view inputs, SlowFast-R50 experienced a 7.2\% drop, and X3D-M achieved the highest overall accuracy of 55.3\% using driver-only inputs. These results highlight that multi-view integration is highly dependent on architectural design. Simple input stacking may introduce representational conflicts, especially in models not built to handle heterogeneous visual information.

Per-class analysis showed that behaviors like head turning and radio interaction remained difficult to classify regardless of input type, emphasizing the challenge of subtle, context-sensitive distractions. Single-pathway models appeared more adaptable to added context than dual-pathway architectures, pointing to the need for design strategies that align well with fusion objectives.

Future work should explore separate processing pipelines for each view, followed by feature-level fusion, to better harness complementary information. Investigating attention-based architectures and applying these techniques to larger, more diverse datasets may further advance the development of effective, context-aware driver monitoring systems.

\section{Acknowledgements}
The authors would like to thank itiger (\url{https://itiger-cluster.github.io/}) for providing access to GPUs for model training.

\section{Declaration of Conflicting Interests}
The authors declared no potential conflicts of interest with respect to the research, authorship, and/or publication of this article.

\end{document}